%% file: Main.tex
\documentclass[letterpaper, 10 pt, conference]{ieeeconf}

\IEEEoverridecommandlockouts
\usepackage{algorithmic}
\usepackage{textcomp}
\usepackage{booktabs}
\usepackage{threeparttable}
\usepackage{pifont}

\usepackage{graphics}
\usepackage{epsfig}
\usepackage{times} 
\usepackage{amsmath,amssymb,amsfonts}
\usepackage[table]{xcolor}
\usepackage[ruled,linesnumbered, noend]{algorithm2e}

\usepackage{cite}
\usepackage{comment}
\usepackage{multirow}
\usepackage{graphicx}
\usepackage{wrapfig}
\usepackage{subfigure}
\usepackage{balance}
\usepackage{wrapfig}
\usepackage{booktabs}  

\usepackage{soul, color}
\usepackage{colortbl}
\definecolor{Gray}{gray}{0.9}

\newcommand{\xmark}{\ding{55}}

\begin{document}

\title{Robust Dynamic Gesture Recognition at Ultra-Long Distances}

\author{Eran Bamani Beeri, Eden Nissinman and Avishai Sintov
\thanks{E. Bamani Beeri, E. Nissinman and A. Sintov are with the School of Mechanical Engineering, Tel-Aviv University, Israel. E-mail: eranbamani@mail.tau.ac.il,  edennissinman@gmail.com, sintov1@tauex.tau.ac.il}
\thanks{This work was supported by the Israel Innovation Authority (grant No. 77857) and the Israel Science Foundation (grant No. 1565/20).} 
}

\maketitle
\thispagestyle{empty}
\pagestyle{empty}

\input{abstract}

\section{Introduction}
\input{Introduction}

\label{sec:introduction}

\section{Methods}

\input{Methods}
\label{sec:method}

\section{Model Evaluation}
\input{Evaluation}

\label{sec:Evaluation}

\section{Conclusions}

\input{Conclusions}

\bibliographystyle{IEEEtran}
\bibliography{ref}

\end{document}

%% file: abstract.tex
\begin{abstract}

Dynamic hand gestures play a crucial role in conveying nonverbal information for Human-Robot Interaction (HRI), eliminating the need for complex interfaces. Current models for dynamic gesture recognition suffer from limitations in effective recognition range, restricting their application to close proximity scenarios. In this letter, we present a novel approach to recognizing dynamic gestures in an ultra-range distance of up to 28 meters, enabling natural, directive communication for guiding robots in both indoor and outdoor environments. Our proposed SlowFast-Transformer (SFT) model effectively integrates the SlowFast architecture with Transformer layers to efficiently process and classify gesture sequences captured at ultra-range distances, overcoming challenges of low resolution and environmental noise. We further introduce a distance-weighted loss function shown to enhance learning and improve model robustness at varying distances. Our model demonstrates significant performance improvement over state-of-the-art gesture recognition frameworks, achieving a recognition accuracy of 95.1\% on a diverse dataset with challenging ultra-range gestures. This enables robots to react appropriately to human commands from a far distance, providing an essential enhancement in HRI, especially in scenarios requiring seamless and natural interaction.

\end{abstract}

%% file: Introduction.tex
Human-robot interaction (HRI) is a rapidly evolving field that aims to enable seamless communication between humans and robots, especially as robots become increasingly integrated into public and private spaces \cite{goodrich2008human, maurtua2017natural}. Intuitive interaction mechanisms are essential to enable non-expert users to effectively communicate their intentions to robots. Hand and arm gestures are natural forms of nonverbal communication that humans use extensively in their daily interactions, making them an ideal medium for HRI \cite{gao2020robust, wang2022hand, urakami2023nonverbal}. Gesture-based interactions reduce the need for complex verbal or physical interfaces, improve user experience, and allow for intuitive and rapid communication with robots, even for users with no technical expertise \cite{wachs2011vision}. With gestures, a user can convey nonverbal and simple commands even from a long distance without the need to shout. For instance, a user may direct robot movements with pointing gestures \cite{bamani2023recognition}.

\begin{figure}
    \centering
    \includegraphics[width=\linewidth]{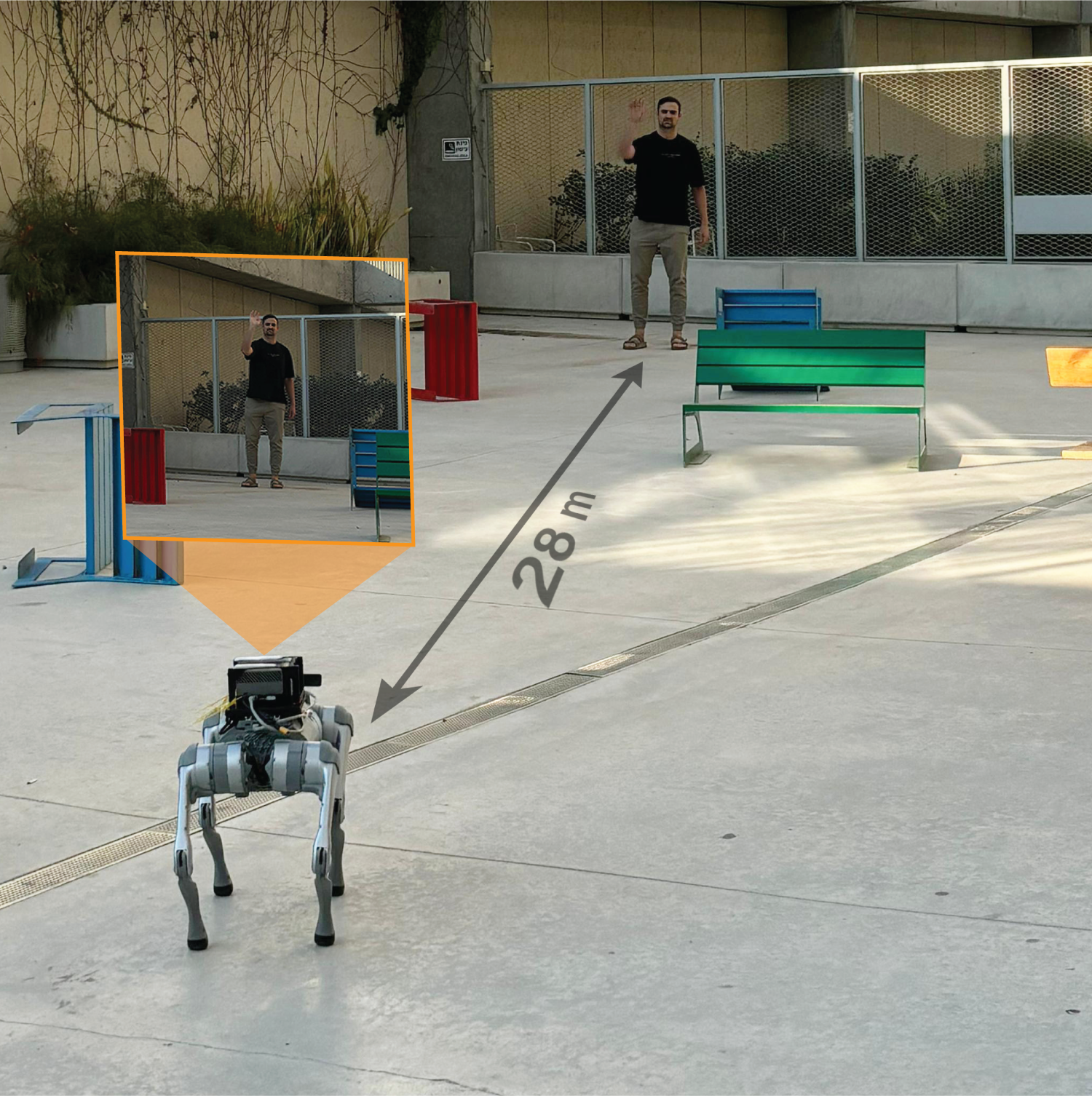}
    \caption{Demonstration of a user instructing a robot to go back by sweeping an open palm forward and backward, from an ultra-range distance. In addition to the low-resolution view of the user's hand, the robot may confuse the dynamic gesture with the static stop gesture.}
    \label{fig:Front}
\end{figure}

Research in the field of gesture recognition has made significant progress, particularly in recognizing static gestures \cite{zhang2020racon, yu2021multi,Yu2022}. However, most gesture recognition approaches are limited to short-range interactions, typically within a few meters \cite{nickel2007visual, wachs2011vision, wang2022hand}. Moreover, gesture recognition methods, often described to be effective in the long range, are typically limited to around seven meters \cite{Zhou9561189,Liang2024}. However, this limitation hinders their applicability in real-world scenarios that demand longer-range HRI. Recognizing gestures from a truly long distance can significantly expand the potential applications of robots in environments such as public spaces, industrial settings, and emergencies, where natural, non-contact interactions are necessary \cite{kim2014non, canal2015gesture}. 
However, one of the key challenges in achieving effective long-range gesture recognition is the degradation of visual information due to factors such as reduced resolution, lighting variations, and occlusions \cite{shen2019overview, pla2018three}. 
In a recent work by the authors, a gesture recognition model was proposed using a simple web camera with an effective ultra-range distance of up to at least 25 meters, addressing some of these challenges \cite{bamani2024ultra}. However, the approach only considers static gestures while dynamic gesture recognition at ultra-range distances has not yet been addressed.

Dynamic gestures provide valuable context for understanding actions and systems, especially when explaining processes that unfold over time \cite{Kang2016}. They are particularly effective for conveying spatial context to a task command \cite{Clough2020}. For instance, a gesture instructing a robot to move backward would usually include an open palm swept back and forth toward it, as demonstrated in Figure \ref{fig:Front}. However, a snapshot of the gesture may be recognized as a static stop instruction if not processed correctly. Several studies have attempted to address dynamic gesture recognition by utilizing various data modalities. RGB-Depth (RGB-D) cameras are often used to recognize dynamic gestures but are limited to indoor environments \cite{xu2015online,ma2018kinect,kabir2019novel, bokstaller2021dynamic,Gao2022}. Approaches using simple RGB cameras were also proposed while shown to function only indoors \cite{Zhou9561189,Yi2018,dos2020dynamic}. Table \ref{tb:sota} summarizes the prominent work on dynamic gesture recognition with visual perception, highlighting the limitations of current approaches, which are typically restricted to indoor environments and short-range interactions (within 7 meters). In it also worth noting on-body sensors in wearable devices that instantly recognize hand gestures \cite{Pyun2023, Tchantchane2023}. While they can offer high recognition rates, wearable devices often require specialized and expensive hardware, limiting their accessibility to occasional users. Additionally, biometric approaches may not generalize well to new users, requiring additional data collection for each individual. On the other hand, visual perception allows for the observation of any user.


\begin{table}[h]
    \centering
    \caption{State-of-the-art comparison of dynamic gesture recognition methods using visual perception}
    \label{tb:sota}
    \begin{tabular}{lcccc}
        \toprule
        Paper & Camera & Range & Indoor & Outdoor \\
        \midrule
        Zhou et al. \cite{Zhou9561189} & RGB & $\leq 7m$ & \checkmark & \xmark \\
        Xu et al. \cite{xu2015online} & RGB-D & $< 1m$ & \checkmark & \xmark \\
        Ma et al. \cite{ma2018kinect} & RGB-D & $\leq 2m$ & \checkmark & \xmark \\
        Kabir et al. \cite{kabir2019novel} & RGB-D & $< 1m$ & \checkmark & \xmark \\
        Bokstaller et al. \cite{bokstaller2021dynamic} & RGB-D & - & \checkmark & \xmark \\
        Gao et al. \cite{Gao2022} & RGB-D & $< 2m$ & \checkmark & \xmark \\
        Yi et al. \cite{Yi2018} & RGB & $\leq 6m$ & \checkmark & \xmark \\
        dos Santos et al. \cite{dos2020dynamic} & RGB & - & \checkmark & \xmark \\
        Tran et al. \cite{tran2019dynamic} & 5$\times$RGB-D & $< 2m$ & \checkmark & \xmark \\
        Wu et al. \cite{wu2016deep} & RGB-D & $< 2m$ & \checkmark & \xmark \\
        \midrule
        Proposed method & RGB & $\leq 28m$ & \checkmark & \checkmark \\
        \bottomrule
    \end{tabular}
\end{table}

While the above methods show feasibility in controlled indoor settings, they often lack scalability and robustness for outdoor environments or ultra-range interactions. Their reliance on specialized hardware, such as a depth cameras, can increase cost and complexity, limiting their practical applicability \cite{ji20123d}. Recent approaches utilizing multimodal data, including RGB, depth and human-pose (i.e., skeleton) information, have demonstrated potential for dynamic gesture recognition \cite{tran2019dynamic, wu2016deep}. However, these methods are often also constrained by limited operational range and the need for specialized equipment. This makes them less suitable for scenarios requiring long-range interactions or deployment in large, open spaces. To the best of the authors' knowledge, no work has addressed the problem of recognizing dynamic gestures in distances farther than seven meters.

%


\begin{figure*}
    \vspace{0.5cm}
    \centering
    \includegraphics[width=\linewidth]{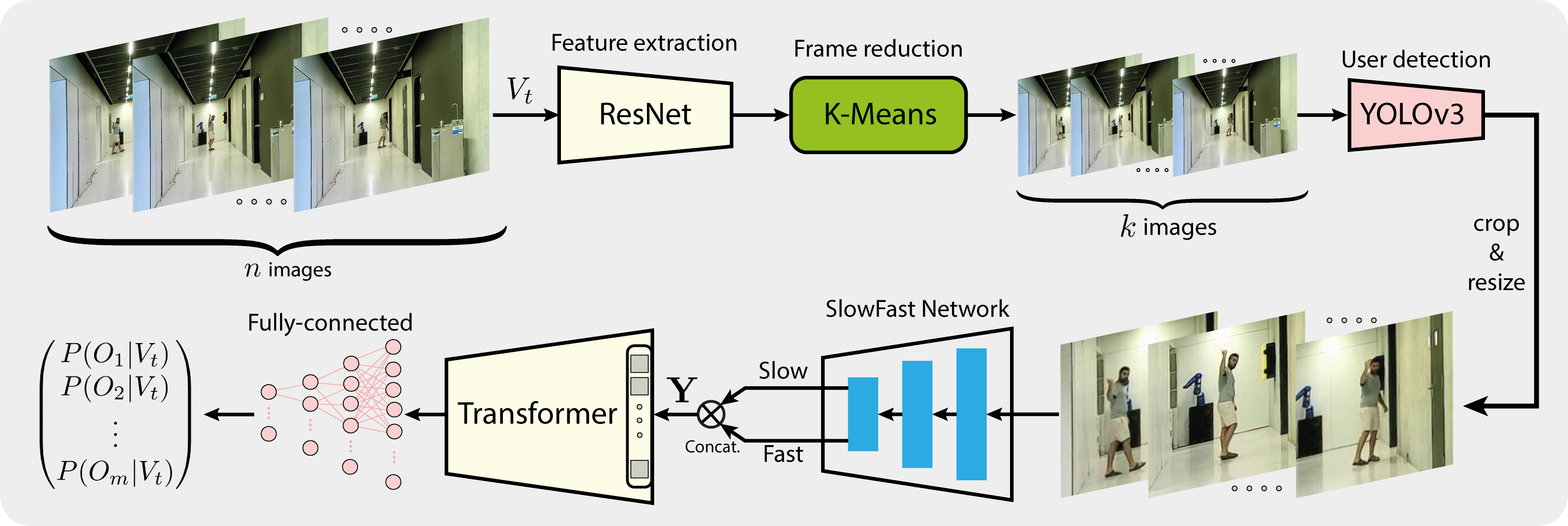}
    \caption{Overview of the proposed SFT framework for dynamic hand gesture recognition. The framework starts with feature extraction using ResNet, followed by frame reduction using K-Means clustering. User detection is performed using YOLOv3, with the output frames resized to $224 \times 224$. The reduced frames are processed through the SlowFast network to capture both slow and fast motion dynamics. The outputted features from the Slow and Fast pathways are concatenated and passed to a Transformer encoder to capture temporal dependencies. Finally, a classification head is employed to acquire the gesture class.}
    \label{fig:scheme}
\end{figure*}


In addition to capturing temporal features of dynamic gestures, the challenge is further compounded by the difficulty of perceiving spatial motions at far distances. In this letter, we address this problem of dynamic gesture recognition at ultra-range distance by only using a simple RGB camera. Specifically, we propose the Slow-Fast-Transformer (SFT) model, illustrated in Figure \ref{fig:scheme}, which integrates the SlowFast architecture \cite{Feichtenhofer} with Transformer encoders \cite{vaswani2017attention}. The integration of SlowFast and Transformer architectures is a key innovation in our model. The SlowFast architecture processes video frames at multiple temporal resolutions, capturing both slow and fast motion dynamics effectively \cite{ xiao2020audiovisual, wu2020multigrid}. By leveraging this with the ability of Transformer encoders to model long-range dependencies in sequential data \cite{grigsby2021long, tucc2020self}, our model can effectively capture the temporal and spatial dynamics of gestures, even at distances up to 28 meters. This integration enables our model to overcome challenges posed by low resolution and environmental variations, which are common in ultra-range gesture recognition scenarios \cite{pla2018three, ren2011robust}. 

To encourage the model to perform better at ultra-range distances, we introduce the Distance-weighted Cross-Entropy (DCE) loss function. This specialized loss function adjusts the contribution of each training sample based on the distance at which the gesture is performed, encouraging the model to better improve its accuracy in recognizing gestures at greater distances. The proposed SFT model trained using the DCE loss yields a robust and generalizable recognition for dynamic gestures in both indoor and outdoor environments. The model addresses key challenges in ultra-range gesture recognition, such as the degradation of visual information due to reduced resolution and lighting variations \cite{gao2024challenges}. Furthermore, we introduce two novel metrics specifically designed to evaluate gesture recognition performance at long distances, including the stability of the recognition over time. Our approach is evaluated using these metrics, along with standard evaluation metrics.

To summarize, the key contributions of this work are:
\begin{itemize} 
\item We propose the novel SlowFast-Transformer (SFT) model, which effectively integrates the strengths of the SlowFast architecture and Transformer encoders to capture both fast and slow temporal dynamics of gestures.
\item We introduce the Distance-weighted Cross-Entropy (DCE) function, a loss function that improves recognition robustness across varying distances, enabling effective ultra-range gesture recognition.
\item Unlike prior work, the proposed SFT model is the first to enable dynamic gesture recognition at ultra-range distances of up to 28 meters, in both indoor and outdoor environments.
\item We provide a comprehensive evaluation of our model against state-of-the-art gesture recognition frameworks, demonstrating superior performance in challenging ultra-range scenarios. The evaluation is conducted using novel metrics specifically designed for ultra-range gesture recognition.
\item The trained models and datasets are publicly available to facilitate further research and development within the community\footnote{To be available upon acceptance for publication.}.
\end{itemize}

By enabling accurate dynamic gesture recognition at ultra-range distances, we significantly expand the potential applications of robots in environments such as public spaces, industrial settings, and emergency situations, where natural, non-contact interactions are necessary. Furthermore, our approach may enhance the ability of robots to interact with multiple users simultaneously, improving their utility in crowded or dynamic environments \cite{bonci2021human, mohd2022multi}. Our method can also be applied in various fields, including search and rescue, drone operations and service robotics. Additionally, its applications may extend to space exploration, law enforcement and entertainment.

%% file: Methods.tex
\subsection{Problem Formulation}
The primary objective of our work is for a robot to accurately recognize a human's dynamic hand gestures at distances of up to 28 meters in diverse environments. Given a set of $m$ gestures $O_1, O_2, \dots, O_m$, where each gesture can be either static or dynamic, the recognition task involves identifying the gesture performed in front of a simple RGB camera. The use of a standard RGB camera ensures that the dataset is applicable in real-world scenarios without the need for specialized hardware. Given an exhibited gesture $O_j$ captured by the camera in a video sequence $V_t=\{I_{t-n+1}, \dots, I_{t}\}$ of $n$ past frames at time $t$ ($I_t$ is a video frame taken at time $t$), we seek to maximize the conditional probability $P(O_j \mid V_t)$. Hence, our formulation aims to find $O_{j^*}$ acquired from the solution of optimization problem
\begin{equation}
    j^* = \underset{j}{\mathrm{argmax}} \; P(O_j \mid V_t), \quad j = 1, \dots, m~.
\end{equation}
Considering a sequence of past images enables the model to account for both the temporal and spatial dynamics of gestures, which is critical for differentiating between gestures that may appear similar when observed in a single frame.



\subsection{Data Collection}

To train a gesture recognition model, a comprehensive dataset of hand gestures is required. A video $V_t$, i.e., sequence of images captured by simple RGB camera, is given showing a user exhibiting a gesture $O_i$ at a distance $d_i\leq28~m$. This yields a dataset $\mathcal{D} = \{(V_i, d_i, o_i)\}_{i=1}^{N}$ of $N$ labeled video sequences, where $o_i\in\{1,\ldots,m\}$ is the gesture index.

To enhance the dataset and improve the model's generalizability, data augmentation techniques are applied to simulate various real-world conditions. These techniques included random cropping, horizontal flipping, rotation, scaling, brightness and contrast adjustments, and synthetic noise addition. These are aimed at simulating different camera positions, lighting conditions, viewing angles, and environmental challenges. The augmented dataset $\tilde{\mathcal{D}} = \{(\tilde{V}_i, d_i, o_i)\}_{i=1}^{M}$ consists of both original and augmented video sequences, such that $M > N$. This enhanced dataset provides diverse training examples, thus improving the model's robustness and performance under real-world variability.


\subsection{Sequence Pre-processing}
\label{sec:preprocess}

A sequence $V_t$ contains $n$ images, some of which may be highly similar, leading to redundancy. Hence, the sequences in $\tilde{\mathcal{D}}$ are pre-processed to reduce the number of frames, resize them, and extract relevant features. Given a video $V_t$, the frames are reduced to $k < n$ representative frames using K-Means clustering. Each frame $\mathbf{I}_j\in V_t$ was processed through a pre-trained ResNet \cite{He} to extract a low-dimensional feature vector $\mathbf{v}_j \in \mathbb{R}^d$. All feature vectors were clustered into $k$ clusters, and the frame corresponding to the feature vector closest to each cluster's centroid was selected as the representative frame. In practice, a representative feature vector is chosen from cluster $C_i=\{v_1,v_2,\ldots\}$ according to
\begin{equation}
    \text{rep}(i) = \arg\min_{j} \|\mathbf{v}_j - \mathbf{C}_i\|_2,
\end{equation}
yielding $k$ representative frames $\{\mathbf{I}_{\text{rep}, 1}, \mathbf{I}_{\text{rep}, 2}, \ldots, \mathbf{I}_{\text{rep}, k}\}$. YOLOv3 \cite{Redmon} was then used to detect the user within each frame and crop the background, enhancing focus particularly when the user was far from the camera. To address cases where the bounding box did not fully capture the human body, the bounding box was extended while maintaining a constant aspect ratio. Specifically, the pixel extension added around the bounding box was $\frac{b}{a}$, where $b$ is the diagonal length of the bounding box, and $a$ is a predefined user-to-image ratio parameter. The resulting cropped image was resized to $224 \times 224$ pixels to ensure uniformity across the dataset, maintaining a consistent size and aspect ratio.

To stabilize the training process, the frames were also normalized to have zero mean and unit variance. In addition, optical flow was computed between consecutive frames to capture motion dynamics, providing information on the direction and magnitude of motion, which is essential for distinguishing between gestures with similar spatial features but different temporal movements. The optical flow data was used as an additional input channel alongside the RGB frames, providing both spatial and temporal information to the model. Dataset $\tilde{\mathcal{D}}$ is updated to include the reduced and enhanced frames.


\subsection{SFT Model Framework}

The proposed SFT model integrates the SlowFast architecture with Transformer encoders to effectively capture the spatial and temporal dynamics of gestures at long ranges. The SlowFast architecture processes video data at multiple temporal resolutions with two distinct pathways. The \textit{Slow} pathway operates at a lower frame rate, capturing slow and steady movements, while the \textit{Fast} pathway operates at a higher frame rate, capturing rapid motions. This dual-pathway design allows the model to efficiently capture both fast and slow components of dynamic gestures. The output features from the Slow and Fast pathways are concatenated into a combined feature representation, denoted by $\mathbf{Y} \in \mathbb{R}^{B \times (c_{\text{slow}} + c_{\text{fast}}) \times T/4 \times H/4 \times W/4}$, where $B$ is the batch size, $c_{\text{slow}}$ and $c_{\text{fast}}$ are the number of channels for the slow and fast pathways, $T$ is the temporal length, and $H$ and $W$ are the height and width of each frame, respectively.

The concatenated features $\mathbf{Y}$ are fed into a Transformer encoder, which leverages the self-attention mechanism to model long-range dependencies across the sequence of frames. The Transformer encoder consists of multiple layers of self-attention and feed-forward networks, allowing the model to focus on different parts of the sequence and effectively capture temporal relationships between frames. The architecture includes convolutional layers, an embedding layer, and four Transformer encoders, enabling the extraction of both spatial and temporal features. The output of the Transformer encoder is temporally pooled and then passed through a global average pooling layer, followed by a fully connected layer with a Softmax activation function to produce the final gesture classification. This classification head ensures that the model can effectively interpret and categorize the dynamic gestures present in the video sequences.

To train the SFT model, we propose the Distance-weighted Cross-Entropy (DCE) loss function, which accounts for the user's distance from the camera. Given a predicted output $\tilde{\mathbf{o}}_i$, target label $\mathbf{o}_i$, and distance $d_i$, the DCE loss is defined by
\begin{equation}
\label{eq:DCE}
\mathcal{L} = \frac{1}{B} \sum_{i=1}^{B} \left[ \text{CE}(\tilde{\mathbf{o}}_i, \mathbf{o}_i) \cdot \left(1 + \frac{\alpha (d_i - b_0)}{b_1 - b_0} \right) \right],
\end{equation}
where $\alpha$ is a distance weighting factor, $b_0$ and $b_1$ are predefined distance thresholds, and $\text{CE}(\cdot)$ is the cross-entropy loss function. This formulation ensures that the model learns to remain robust across varying distances, making it suitable for real-world deployment in diverse environments. Overall, the combination of the SlowFast architecture and a Transformer encoder allows the SFT model to effectively capture both spatial and temporal features, aiming for robust recognition of dynamic gestures even at ultra-range distances.

%% file: Evaluation.tex
In this section, we evaluate the proposed SFT framework for recognizing dynamic gestures used to naturally guide a robot. All computations were conducted on a Linux Ubuntu 18.04 LTS system equipped with an Intel Xeon Gold 6230R CPU (20 cores at 2.1 GHz) and four NVIDIA GeForce RTX 2080TI GPUs, each with 11 GB of RAM. Hyperparameter tuning was performed using Ray-Tune \cite{Liaw2018} to all models. 


\subsection{Gestures}

\begin{figure}
    \centering
    \includegraphics[width=\linewidth]{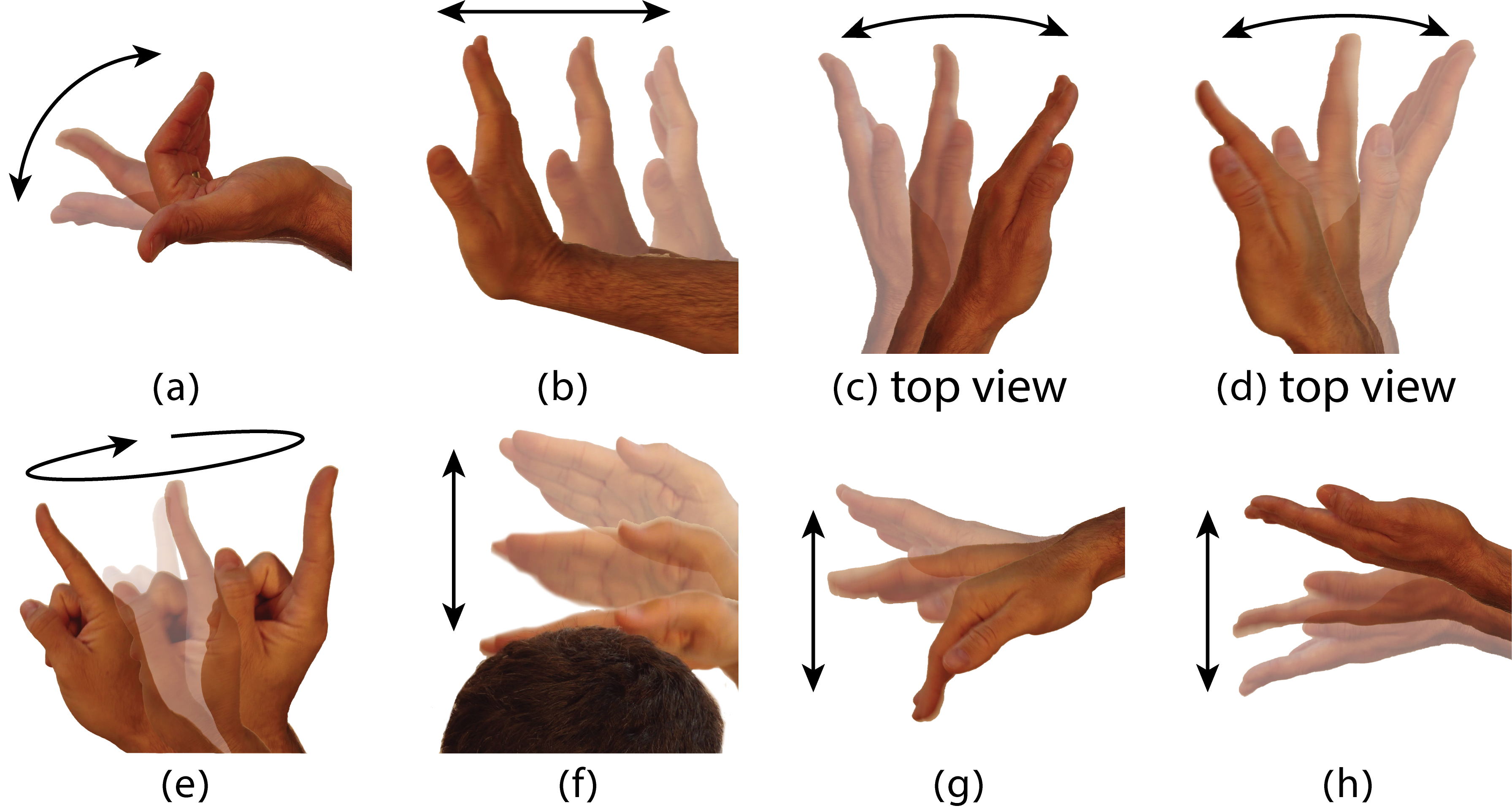}   
    \caption{The eight dynamic gestures used in the analysis include: (a) beckoning, (b) go-back, (c) move-right, (d) move-left, (e) turn-around, (f) follow-me, (g) go-down and (h) go-up.}
    \label{fig:human_gestures}
\end{figure}

The evaluation focuses on $m=13$ distinct gesture classes. Of the 13, eight are dynamic, seen in Figure \ref{fig:human_gestures}, and include: \textit{go-back} with a forward and backward motion of an open hand, palm facing outward; \textit{go-up} with an upward motion of an open hand, palm facing upward; \textit{go-down} with a downward motion of an open hand, palm facing downward; \textit{move-right} with a horizontal sweeping motion of the open hand, palm facing right; \textit{move-left} with horizontal sweeping motion of the open hand, palm facing left; \textit{turn-around} with a circular motion of the index finger, pointing upwards; \textit{beckoning} where the palm is facing upward and the fingers are rhythmically flexed and extended; \textit{follow-me} where the open palm is repeatedly tapped on the head of the user. 
Another four static gestures are included: \textit{pointing}, \textit{thumbs-up}, \textit{thumbs-down}, and \textit{stop}. These gestures were chosen due to their mainstream usage and to challenge our model. That is, confusion between static and dynamic gestures may occur such as in turn-around vs. pointing, go-back vs. stop, and go-up vs. beckoning. All gestures can be performed with either the left or right arms. The last class is the \textit{null} which represents the absence of any exhibited gesture, in which the user can perform any unrelated activities.

\subsection{Dataset}

Dataset $\mathcal{D}$ was collected by recording video samples of the $m=13$ gestures at a distance range of $d\in[2,28]$. Gestures were exhibited in diverse environments, including both indoor and outdoor settings, varying lighting, and diverse backgrounds. In addition, 16 different participants contributed gesture data to ensure variability in gesture performance. Each participant performed each gesture multiple times at different measured distances. The data collection process involved using a standard RGB camera with a resolution of $640 \times 480$ pixels taken at $21$ frames per second. Each gesture was manually annotated according to its class $o_i$ and the distance $d_i$ at which it was performed, resulting in labeled data suitable for supervised learning. This annotation allowed for a detailed analysis of model performance across varying distances, as well as the development of distance-specific recognition strategies. The collected dataset $\mathcal{D}$ yielded $N = 3,240$ video samples of hand gestures, each lasting 4 seconds with up to $n = 84$ frames. After feature extraction as described in Section \ref{sec:preprocess}, the number of frames was reduced to $k = 8$ per video. With further augmentation, dataset $\tilde{\mathcal{D}}$ includes $M=4,790$ samples. An additional test set of $K=458$ labeled and processed videos was recorded in distinct environments to evaluate the model's performance.

\subsection{Comparative Evaluation}

We now evaluate the proposed SFT model compared to existing state-of-the-art gesture recognition frameworks. The SFT model was trained using the Adam optimizer with an initial learning rate of 0.001. The learning rate was gradually reduced using a cosine annealing schedule to ensure stable convergence. The model was trained for 100 epochs with a batch size of $B=16$, using the DCE loss function \eqref{eq:DCE}. To prevent overfitting, dropout was applied at the output of the Transformer encoder, and L2 regularization was used during training. Early stopping was also employed, with the validation loss monitored to determine the stopping point. 

The SFT model is compared to the: Swin Transformer \cite{Liu} which uses shifted windows for efficient spatial feature extraction; ViViT \cite{Arnab}, a Transformer for video analysis capturing spatial and temporal relationships; TimeSformer \cite{gberta_2021_ICML} which separates temporal and spatial attention; MViT \cite{fan2021multiscale}, a multiscale Transformer for spatiotemporal resolutions; I3D \cite{carreira2017quo}, a 3D convolutional network for spatiotemporal learning; and, X3D \cite{Feichtenhofer2}, an efficient 3D convolutional model balancing speed and accuracy. 

The models were trained and evaluated on the same datasets to ensure a fair comparison. We employ five key metrics: recognition success rate, Mean Average Precision (mAP), $F_1$ Score, Distance-Weighted Accuracy (DWA), and Gesture Stability Score (GSS). While the three former ones are common metrics, we propose to include the latter two for a comprehensive evaluation across distances and their consistency over time. The DWA emphasizes correct classifications of gestures performed at greater distances, reflecting the model's robustness. DWA is defined as:
\begin{equation}
\text{DWA} = \frac{1}{K} \sum_{i=1}^{K} \mathbb{I}(\tilde{o}_i = o_i)w_i ,
\end{equation}
where 
, $w_i = 1 + \beta \frac{d_i - d_{\text{min}}}{d_{\text{max}} - d_{\text{min}}}$, $d_i$ is the distance at which gesture $i$ was performed, $d_{\text{min}}=2~m$ and $d_{\text{max}}=28~m$ are the minimum and maximum distances, respectively, $\beta=1.6$ controls the weight for long distances, and $\mathbb{I}(\tilde{o}_i = o_i)$ equals to 1 if the predicted label $\tilde{o}_i$ matches the true label $o_i$, otherwise 0. While DWA evaluates the success rate with an emphasis on the farther cases, GSS measures the stability of predictions across the frames, ensuring consistent recognition. GSS is defined by
\begin{equation}
\text{GSS} = \frac{1}{K} \sum_{i=1}^{K} \left( \frac{1}{n_i} \sum_{j=n}^{n_i} \mathbb{I}(\tilde{o}_{i, j} = o_i) \right),
\end{equation}
where $n_i>n$ is the number of frames in video $V_i$, $\tilde{o}_{i, j}$ is the predicted label for sequence $\{I_{j-n+1},\ldots,I_j\}$ in video $V_i$, and $o_i$ is the true label for video $V_i$. A GSS value closer to one indicates a more stable prediction along the video. mAP provides an average measure of precision across all gesture classes, ensuring a balanced evaluation of the model's capability. $F_1$ Score is the harmonic mean of precision and recall, useful for evaluating models where both false positives and false negatives are impactful. These metrics provide a comprehensive evaluation of the SFT model, focusing on accuracy, precision, and stability. Table \ref{tab:comparison} presents the comparative results after cross validation, demonstrating the superior performance of the SFT across all evaluation metrics, particularly in terms of recognition success rate.

\begin{table*}[h]
    \centering
    \caption{Evaluation results for different dynamic gesture recognition models}
    \label{tab:comparison}
    \begin{tabular}{lcccccc}
        \toprule
        \multirow{1}{*}{Model} && Success rate (\%) & DWA & GSS & $F_1$ score & mAP (\%) \\
        \midrule
        Swin &\cite{Liu} & 80.5 & 0.84 & 0.85 & 0.83 & 78.1 \\
        ViViT &\cite{Arnab} & 78.3 & 0.82 & 0.84 & 0.80 & 77.5 \\
        TimeSformer &\cite{gberta_2021_ICML} & 83.4 & 0.85 & 0.87 & 0.85 & 81.3 \\
        MViT &\cite{fan2021multiscale} & 87.9 & 0.88 & 0.90 & 0.89 & 85.1 \\
        I3D &\cite{carreira2017quo} & 84.3 & 0.86 & 0.88 & 0.85 & 82.4 \\
        X3D &\cite{Feichtenhofer2} & 86.2 & 0.87 & 0.89 & 0.87 & 78.8 \\
        SFT && \cellcolor[HTML]{C0C0C0}95.1 & \cellcolor[HTML]{C0C0C0}0.89 & \cellcolor[HTML]{C0C0C0}0.94 & \cellcolor[HTML]{C0C0C0}0.94 & \cellcolor[HTML]{C0C0C0}93.2 \\
        \bottomrule
    \end{tabular}%
\end{table*}



\subsection{SFT analysis}

\begin{figure}[htbp]
\centering
\includegraphics[width=\linewidth]{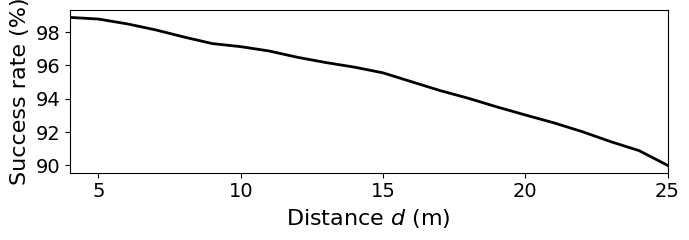}
\caption{Gesture recognition success rate of the SFT model with regard to the distance $d$ of the user from the camera.}
\label{fig:performance_vs_distance}
\end{figure}

\begin{figure}[htbp]
    \centering
    \includegraphics[width=\linewidth]{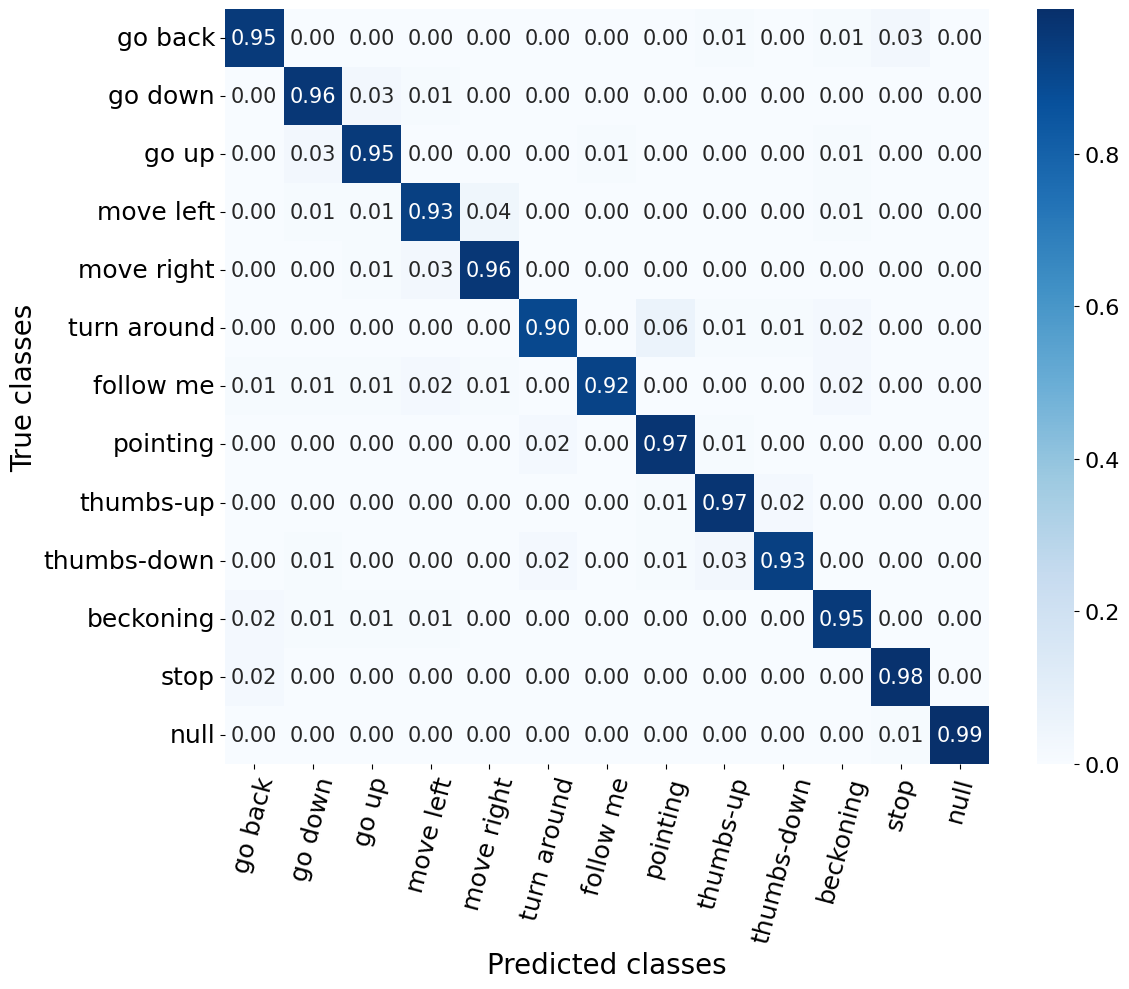} 
    \caption{Confusion matrix for the gesture classification with the SFT model across 13 gesture classes.}
    \label{fig:confmat}
\end{figure}

We further analyze the performance of the proposed SFT model. Figure \ref{fig:performance_vs_distance} presents the gesture recognition success rate with respect to the distance $d$ between the user and the camera. The success rate gradually decreases but remains relatively high in the desired range of up to 28 meters. This demonstrates the robustness of the SFT model in recognizing gestures at ultra-range, though performance diminishes as distance increases due to factors such as reduced resolution and increased visual noise. Figure \ref{fig:confmat} presents the confusion matrix for the SFT model over the test data and for all 13 gesture classes. 
While some gestures may be visually similar, the temporal embedding employed by the SFT effectively distinguishes between them, reducing the likelihood of misinterpretation.


\begin{figure}[htbp]
\centering
\includegraphics[width=\linewidth]{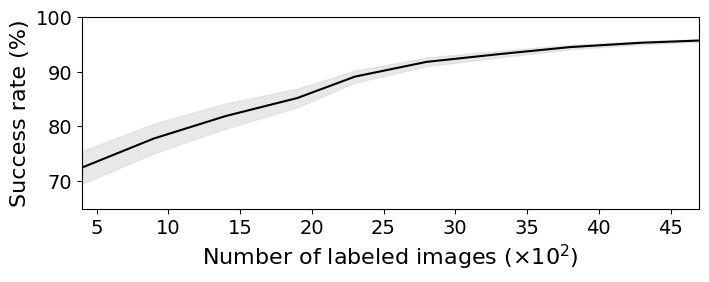}
\caption{Gesture recognition success rate of the SFT model with regard to the number of labeled training images.}
\label{fig:performance_vs_data}
\end{figure}

To understand the impact of the amount of training data on the success rate of gesture recognition, we analyzed the SFT model's performance with a varying number of labeled images. Figure \ref{fig:performance_vs_data} shows how the model's average success rate improves as more labeled images are utilized. For each number of labeled images, the SFT model was trained 10 times with different parts of the dataset, yielding an average success rate. The success rate of the model increases significantly, reaching 95.7\% with the full dataset of 4,790 images. The gradual improvement with increased data demonstrates the model's ability to learn complex gesture patterns effectively as more labeled examples are introduced but also indicates that beyond a certain point, adding more data does not yield substantial gain. 

\begin{figure}[htbp]
\centering
\includegraphics[width=\linewidth]{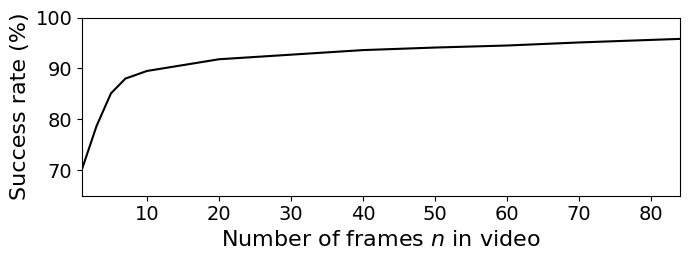}
\caption{Gesture recognition success rate of the SFT model with regard to the number of frames $n$ in a video.}
\label{fig:Success_Num_Frames}
\end{figure}

The above results were acquired with a video window length of up to $n=84$ frames. Hence, we now evaluate the performance of the SFT with respect to the window length. The SFT model was trained multiple times on video sequences $V_t$ of varying lengths $n$, and each trained model's performance was evaluated on test sequences of similar lengths. Figure \ref{fig:Success_Num_Frames} shows the recognition success rate of the SFT with regard to the number of frames $n$ in a video sample. First, with only one image of the gesture (i.e., $n=1$), a poor recognition success rate of $70.3\%$ is achieved, due to the lack of dynamic information. As more frames are added to $V_t$, more dynamic information is included and the recognition improves. A longer video encapsulates more dynamic information on the exhibited gesture, yielding better recognition accuracy. However, increasing $n$ may affect the time sensitivity and frequency of the recognition in real time. Nevertheless, increasing the sequence length beyond $n=84$ frames offers diminishing returns in terms of accuracy.

\begin{table*}[]
    \centering
    \caption{Ablation study results for SFT components}
    \label{tab:Ablation}
    \begin{tabular}{lccccc}
        \toprule
        Model Variant  & Success rate (\%)  & DWA & GSS & $F_1$ score & mAP (\%) \\
        \midrule
        SFT w/o Slow pathway    & 87.2 & 0.85 & 0.87 & 0.86 & 82.3 \\
        SFT w/o Fast pathway    & 89.5 & 0.87 & 0.89 & 0.88 & 85.5 \\
        SFT w/o Transformer     & 86.5 & 0.84 & 0.86 & 0.84 & 81.0 \\
        SFT w/o DCE             & 88.5 & 0.86 & 0.88 & 0.89 & 84.7 \\
        SFT w/o Temporal pooling       & 90.1 & 0.88 & 0.91 & 0.90 & 87.8 \\
        Full SFT model & \cellcolor[HTML]{C0C0C0}95.1 & \cellcolor[HTML]{C0C0C0}0.89 & \cellcolor[HTML]{C0C0C0}0.94 & \cellcolor[HTML]{C0C0C0}0.94 & \cellcolor[HTML]{C0C0C0}93.2 \\
        \bottomrule
    \end{tabular}%
\end{table*}

An ablation study was also conducted to evaluate the contribution of each component of the SFT model to the performance. The study involved training and evaluating the model while removing a single component in each attempt, such as the slow pathway, fast pathway, Transformer encoder, DCE loss, and temporal pooling. The results are summarized in Table \ref{tab:Ablation}, which demonstrates the importance of each component to the overall performance of the model. The full SFT model achieved the highest performance across all metrics, showcasing the importance of all proposed components. For instance, removing the DCE loss and training with standard cross-entropy leads to a decreased focus on farther samples, resulting in reduced accuracy. Similarly, the transformer encoder is shown to be crucial to handle the temporal data. Generally, removing individual components led to a noticeable decline in performance, highlighting the contribution of each to the effectiveness of the SFT model.


\subsection{Sequence Analysis and Varying Conditions}

To evaluate the robustness of the proposed SFT model, we conducted several experiments involving complex gesture sequences and tested the model under varying environmental conditions. In the first experiment, users performed complex gesture sequences consecutively simulating the guidance of a robot. The sequences included various combinations of the 13 gesture classes. A total of 50 unique gesture sequences, each composed of 3 to 5 individual gestures, were used to ensure comprehensive testing. 
Table \ref{tab:combined_results} presents the sequence accuracy, where a sequence is considered successful only if all gestures within it are correctly classified. The high accuracy indicates that the model effectively interpreted and executed the complex gesture sequences. 

Additionally, we evaluated the model's recognition accuracy under varying lighting conditions, including indoor controlled lighting, outdoor bright sunlight, and overcast outdoor settings. The recognition success rates are seen in Table \ref{tab:combined_results}. The high accuracy in controlled indoor environments demonstrates the model's ability to recognize gestures under optimal conditions, while the slightly lower accuracy in outdoor environments reflects the challenges posed by changing lighting and environmental noise. Nevertheless, these results demonstrate the model's robustness in adapting to different environmental conditions, ensuring reliable performance in diverse real-world scenarios.

\begin{table}[]
\centering
\caption{Results of Sequence Analysis and Varying Conditions}
\begin{tabular}{lc}
    \toprule
    Metric & Success rate (\%) \\
    \midrule
    Sequence accuracy               & 92.3\% \\
    Controlled lighting (Indoor)    & 95.7\% \\
    Bright sunlight (Outdoor)       & 91.5\% \\
    Overcast (Outdoor)              & 89.8\% \\
    \bottomrule
\end{tabular}
\label{tab:combined_results}
\end{table}



%% file: Conclusions.tex
In this work, we proposed the SFT model for recognizing dynamic hand gestures at ultra-range distances of up to 28 meters with only an RGB camera, addressing the challenges of low resolution and environmental noise in both indoor and outdoor environments. Our approach effectively integrates the SlowFast architecture with Transformer encoders, capturing both fast and slow temporal dynamics to enhance recognition accuracy across varying conditions. Additionally, the introduction of the DCE loss function further improved the model's robustness at greater distances. Experimental results demonstrated that the SFT model achieves a high recognition accuracy of 95.1\% on a diverse dataset, outperforming state-of-the-art gesture recognition frameworks. 

The model's capability to recognize gestures accurately over long distances significantly expands the applicability of robots in real-world scenarios, such as public spaces, industrial environments, and emergency situations, where seamless and natural interaction is essential. Future work may focus on expanding the gesture vocabulary, including non-intentional body language and facial expressions, optimizing real-time performance, and further enhancing the model's robustness under challenging conditions, such as dynamic backgrounds, varying lighting, and multiple users in the scene.

